\documentclass[sigconf, screen=true, review=false, printacmref=false, printccs=false, printfolios=false]{acmart}
\usepackage{comment}
\graphicspath{plots}
\usepackage{multirow}
\setcopyright{none}
\usepackage{paralist}
\usepackage{booktabs} 
\usepackage{epstopdf}
\usepackage{comment}
\usepackage{tabularx}
\usepackage{subfigure}
\usepackage{enumitem}

\DeclareMathAlphabet\mathbfcal{OMS}{cmsy}{b}{n}

\usepackage{xcolor}
\definecolor{thedarkblue}{RGB}{0,0,120} 
\definecolor{theblue}{rgb}{0,0.08,0.45} 
\definecolor{darkblue}{rgb}{0,0.08,180}
\colorlet{TufteRed}{red!80!black}

\definecolor{theblue}{RGB}{0,0,180}
\colorlet{thered}{TufteRed}
\usepackage{microtype}
\usepackage{balance}
\usepackage{amsmath,amssymb,amsthm}

\usepackage{microtype}
\usepackage{balance}
\usepackage{setspace}
\graphicspath{{./}{./graphics/}}
\newcolumntype{H}{>{\setbox0=\hbox\bgroup}c<{\egroup}@{}}

\newcolumntype{R}[1]{>{\RaggedLeft\arraybackslash}} 
\newcolumntype{L}[1]{>{\RaggedRight\arraybackslash}}

\newcommand{\eg}{\emph{e.g.}}
\newcommand{\ie}{\emph{i.e.}}

\newtheorem{Definition}{\hspace{-1em}\bfseries{Definition}}

\setcopyright{rightsretained}
\usepackage{nicefrac}
\usepackage{numprint} 
\definecolor{thelightblue}{RGB}{0,191,255}

\usepackage{amsmath,lipsum}

\usepackage{epstopdf}

\makeatletter 
\def\vcdots{\vbox{\baselineskip4\p@ \lineskiplimit\z@
\kern3\p@\hbox{.}\hbox{.}\hbox{.}\kern3\p@}}
\makeatother

\usepackage{blkarray}
\usepackage{bigstrut, booktabs}

\makeatletter
\renewcommand*\env@matrix[1][*\c@MaxMatrixCols c]{
\hskip -\arraycolsep
\let\@ifnextchar\new@ifnextchar
\array{#1}}
\makeatother

\usepackage{booktabs} 
\usepackage{enumitem}
\usepackage{subfigure}
\usepackage{rotating}
\usepackage{multirow}
\usepackage{comment}

\usepackage{color}
\usepackage{xcolor}
\definecolor{mydarkblue}{RGB}{0, 20, 159} 
\definecolor{mydarkblue}{rgb}{0,0.08,0.45} 

\DeclareSymbolFont{cmbrightop}{OT1}{cmbr}{m}{n}
\DeclareMathSymbol{\sfPsi}{\mathalpha}{cmbrightop}{9}

\usepackage{tikz}
\usepackage{verbatim}
\usetikzlibrary{arrows}
\usetikzlibrary{shapes,snakes}
\usetikzlibrary{decorations.pathmorphing} 
\usetikzlibrary{fit}					
\usetikzlibrary{backgrounds}	

\definecolor{gray}{RGB}{150,150,150}
\definecolor{theblue}{RGB}{0, 20, 159} 

\definecolor{myyellow}{RGB}{255,255,204}
\definecolor{myred}{RGB}{255,204,204}
\definecolor{myblue}{RGB}{0,200,255}
\definecolor{mygreen}{RGB}{80,220,80}

\usepackage{balance}

\usepackage{epsfig}

\usepackage{graphicx}

\usepackage{amssymb}
\usepackage{amsmath}
\usepackage{amsfonts}

\usepackage{tabularx}
\usepackage{booktabs}
\usepackage{relsize}

\usepackage{numprint} 

\usepackage{setspace}
\graphicspath{{./}{./graphics/}}
\newcolumntype{H}{>{\setbox0=\hbox\bgroup}c<{\egroup}@{}}

\usepackage{algorithm}
\usepackage{algorithmicx}
\usepackage{algpseudocode}
\algblockdefx[parallel]{ParFor}{EndPar}[1][]{$\textbf{parallel for}$ #1 $\textbf{do}$}{$\textbf{end parallel}$}
\algrenewcommand{\alglinenumber}[1]{\fontsize{6.5}{7}\selectfont#1}
\algtext*{EndPar}

\algblockdefx[parallel]{parfor}{endpar}[1][]{$\textbf{parallel for}$ #1 $\textbf{do}$}{$\textbf{end parallel}$}
\algrenewcommand{\alglinenumber}[1]{\scriptsize#1:}

\algblockdefx[parallel]{parallelfor}{parallelend}
[1][]{\textbf{parallel for} #1}
{\textbf{end parallel}}

\usepackage{nicefrac}

\algnotext{EndIf}
\algnotext{EndProcedure}

\newtheorem{Result}{\bfseries{Conclusion}}
\usepackage[notheorems]{rr-math}

\begin{document}

\title[From Closing Triangles to Closing Higher-Order Motifs]
{Higher-Order Ranking and Link Prediction: \\From Closing Triangles to Closing Higher-Order Motifs}

\author{Ryan A. Rossi}
\affiliation{
\institution{Adobe Research}
}
\email{rrossi@adobe.com}

\author{Anup Rao}
\affiliation{
\institution{Adobe Research}
}
\email{anuprao@adobe.com}

\author{Sungchul Kim}
\affiliation{
\institution{Adobe Research}
}
\email{sukim@adobe.com}

\author{Eunyee Koh}
\affiliation{
\institution{Adobe Research}
}
\email{eunyee@adobe.com}

\author{Nesreen K. Ahmed}
\affiliation{
\institution{Intel Labs}
}
\email{nesreen.k.ahmed@intel.com}

\author{Gang Wu}
\affiliation{
\institution{Adobe Research}
}
\email{gang.wu@adobe.com}
\email{}

\renewcommand{\shortauthors}{R.~A.~Rossi et al.}

\begin{abstract}
In this paper, we introduce the notion of \emph{motif closure} and describe higher-order ranking and link prediction methods based on the notion of \emph{closing higher-order network motifs}. The methods are fast and efficient for real-time ranking and link prediction-based applications such as web search, online advertising, and recommendation. In such applications, real-time performance is critical. The proposed methods do not require any explicit training data, nor do they derive an embedding from the graph data, or perform any explicit learning. Most existing methods with the above desired properties are all based on closing triangles (common neighbors, Jaccard similarity, and the ilk). In this work, we investigate higher-order network motifs and develop techniques based on the notion of closing higher-order motifs that move beyond closing simple triangles. All methods described in this work are fast with a runtime that is sublinear in the number of nodes. The experimental results indicate the importance of closing higher-order motifs for ranking and link prediction applications. Finally, the proposed notion of higher-order motif closure can serve as a basis for studying and developing better ranking and link prediction methods.
\end{abstract}




\maketitle

\section{Introduction} \label{sec:intro}
Link prediction generally refers to predicting the existence of edges (node pairs) in $G$ such that the predicted edges (node pairs) are not in the original edge set $E$ of $G$.
The goal of this task may be to predict future links at time $t+1$ or to simply predict links that were not observed (\eg, to improve the quality of downstream tasks)~\cite{rossi12jair}.
Notice that nearly all link prediction methods first compute a weight $W_{ij}=f(i,j)$ between node $i$ and $j$ and then use $W_{ij}$ to decide whether to predict a link $(i,j)$ or not.
We denote the task of estimating a weight $W_{ij}=f(i,j)$ between node $i$ and $j$ as \emph{link weighting} or \emph{link strength estimation}.
The weights are then used to derive a ranking of potential links.
The potential links may refer to items $j$ that a user $i$ is likely to purchase, or songs that a user is likely to prefer, and so on.
In this work, we focus on fast and efficient methods for computing link weights based on closing higher-order network motifs.
Such weights based on higher-order motif closures can then be used for ranking-based applications (such as recommender systems and the ilk).

Ranking is a key component of many real-world applications such as web search, online advertising, and recommendation~\cite{manning2010introduction}.
In these applications, real-time performance is critical, \eg, in web search users expect an answer to their query in the order of a few hundred milliseconds~\cite{deshpande2004item,berry2005understanding}.
This makes it impossible to learn a complex ranking function.
Instead, there are usually two components to such a system.
In the first component, a \emph{fast online approach} is used to identify the top-$k$ most relevant results in real-time (where $k$ is typically small), which are then displayed to the user.
In the second component, a more accurate but computationally expensive model is trained to improve the initial ranking.
The ranking learned from the model can be used directly or combined with simpler approaches to obtain a final re-ranking of the web pages (or items).
In this work, we primarily focus on the first component.

\begin{figure}[t!]
\vspace{4mm}
\centering
\includegraphics[width=1.0\linewidth]{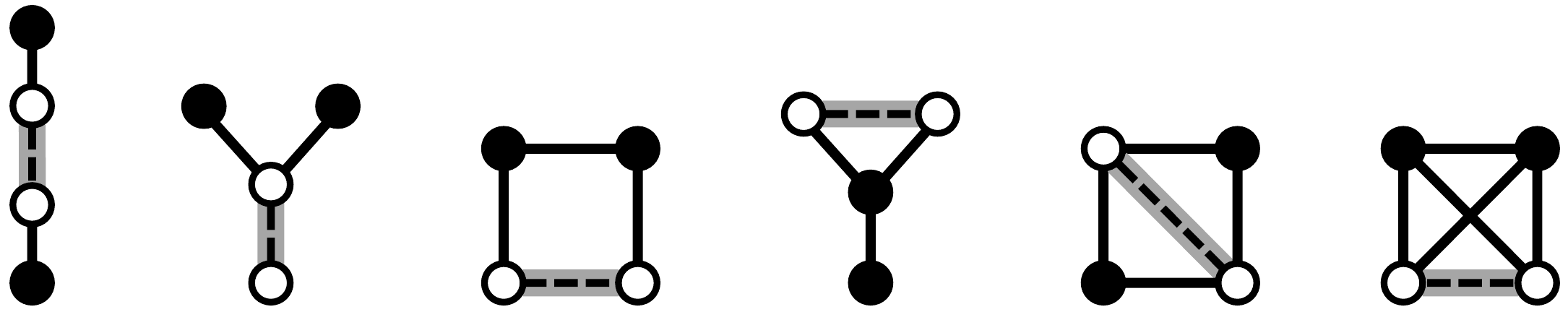}
\vspace{-5mm}
\caption{Higher-Order Motif Closures. 
The unshaded/white nodes are node $i$ and $j$.
Given a node pair $(i,j) \not\in E$ (unshaded/white nodes) and any motif/induced subgraph $H$, the ``edge'' between $i$ and $j$ (dotted gray line) is said to close an instance $F$ of $H$ if the edge $(i,j)$ were to actually exist in $G$.
}
\label{fig:network-motif-operators}
\vspace{-0mm}
\end{figure}

Ranking and link prediction~\cite{deshpande2004item} are important fundamental problems with many applications including recommendation of items~\cite{deshpande2004item}, friends~\cite{liben2007link}, web pages~\cite{manning2010introduction}, among many others~\cite{rossi12jair,manning2010introduction}.
Common neighbors and approaches based on common neighbors such as Jaccard similarity are known to be strong baselines that are hard to beat in practice~\cite{tsitsulin2018verse}.
These baselines are all fundamentally based on the notion of ``closing triangles''~\cite{rossi12jair,pgd}.
They are both simple and fast for ranking in an online real-time fashion.
In this work, we investigate higher-order network motifs and develop ranking techniques based on the notion of \emph{closing higher-order motifs} (Definition~\ref{def:motif-closure}) that move beyond ``closing'' simple triangles.

\begin{table*}[h!]
\centering
\setlength{\tabcolsep}{3.7pt}
\renewcommand{\arraystretch}{1.2}
\scriptsize
\caption{
Mean average precision (MAP) results for ranking (and prediction) methods based on closing higher-order motifs.
}
\vspace{-3mm}
\label{table:results-MAP-10perc}
\small
\fontsize{8.0}{9.0}\selectfont
\npdecimalsign{.}
\nprounddigits{3}
\begin{tabularx}{1.0\linewidth}{@{}
rccccccc c c cccccccc@{}}
\toprule
&
\multicolumn{1}{l}{\rotatebox{80}{\textsf{bn-mouse}}}  &
\multicolumn{1}{l}{\rotatebox{80}{\textsf{bio-DM-LC}}}  &
\multicolumn{1}{l}{\rotatebox{80}{\textsf{bio-CE-HT}}}  &
\multicolumn{1}{l}{\rotatebox{80}{\textsf{bio-DM-HT}}}  &
\multicolumn{1}{l}{\rotatebox{80}{\textsf{ia-reality}}}  &
\multicolumn{1}{l}{\rotatebox{80}{\textsf{web-polblogs}}}  &
\multicolumn{1}{l}{\rotatebox{80}{\textsf{biogrid-worm}}}  &
\multicolumn{1}{l}{\rotatebox{80}{\textsf{biogrid-plant}}}  &
\multicolumn{1}{l}{\rotatebox{80}{\textsf{biogrid-yeast}}}  &
\multicolumn{1}{l}{\rotatebox{80}{\textsf{email-dnc-corec.}}}  &
\multicolumn{1}{l}{\rotatebox{80}{\textsf{soc-advogato}}}  &
\multicolumn{1}{l}{\rotatebox{80}{\textsf{econ-wm1}}}  &
\multicolumn{1}{l}{\rotatebox{80}{\textsf{bn-macaque-rhe.}}}  &
\multicolumn{1}{l}{\rotatebox{80}{\textsf{road-minnesota}}}  &
\multicolumn{1}{l}{\rotatebox{80}{\textsf{soc-fb-messages}}}  &
\multicolumn{1}{l}{\rotatebox{80}{\textsf{email-EU}}}  &
\multicolumn{1}{l}{\rotatebox{80}{\textsf{email-univ}}}  \\

\midrule
\textbf{4-path}  &   0.829  &   0.687  &   0.607  &   0.594  &   0.649  &   0.778  &   0.865  &   0.729  &   0.893  &   0.873  &   0.914  &   0.788  &   0.942  &   0.326  &   0.844  &   0.854  &   0.707  \\
\textbf{4-star}  &   0.880  &   0.787  &   0.595  &   0.696  &   \textbf{0.922}  &   0.814  &   0.895  &   0.861  &   0.840  &   0.813  &   0.889  &   0.688  &   \textbf{0.961}  &   0.388  &   0.807  &   \textbf{0.972}  &   0.695  \\
\textbf{4-cycle}  &   \textbf{0.881}  &   \textbf{0.958}  &   \textbf{0.651}  &   \textbf{0.926}  &   0.827  &   0.885  &   \textbf{0.908}  &   \textbf{0.935}  &   0.927  &   0.957  &   0.930  &   0.900  &   0.773  &   \textbf{0.950}  &   0.870  &   0.902  &   0.847  \\
\textbf{4-tailed-triangle}  &   0.804  &   0.612  &   0.570  &   0.752  &   0.773  &   0.663  &   0.773  &   0.681  &   0.689  &   0.779  &   0.600  &   0.496  &   0.530  &   0.834  &   0.722  &   0.937  &   0.582  \\
\textbf{4-chordal-cycle}  &   0.801  &   0.837  &   0.598  &   0.842  &   0.312  &   \textbf{0.966}  &   0.840  &   0.854  &   \textbf{0.977}  &   0.996  &   \textbf{0.986}  &   0.947  &   0.750  &   0.939  &   0.935  &   0.782  &   0.969  \\
\textbf{4-clique}  &   0.804  &   0.838  &   0.595  &   0.843  &   0.293  &   0.963  &   0.842  &   0.847  &   0.972  &   \textbf{0.997}  &   \textbf{0.986}  &   \textbf{0.965}  &   0.759  &   0.939  &   \textbf{0.960}  &   0.798  &   \textbf{0.982}  \\

\midrule
\textbf{CN}  &   0.705  &   0.872  &   0.613  &   0.839  &   0.422  &   0.814  &   0.833  &   0.897  &   0.839  &   0.960  &   0.949  &   0.852  &   0.342  &   0.945  &   0.790  &   0.890  &   0.941  \\

\textbf{Jaccard Sim.}  &   0.705  &   0.873  &   0.618  &   0.841  &   0.537  &   0.933  &   0.853  &   0.918  &   0.955  &   \textbf{0.997}  &   0.973  &   0.918  &   0.764  &   0.944  &   0.841  &   0.933  &   0.949  \\

\textbf{Adamic/Adar}  &   0.705  &   0.883  &   0.621  &   0.842  &   0.549  &   0.940  &   0.856  &   0.920  &   0.959  &   \textbf{0.997}  &   0.976  &   0.919  &   0.777  &   0.945  &   0.848  &   0.935  &   0.953  \\

\bottomrule
\end{tabularx}
\npnoround
\vspace{-0mm}
\end{table*}

While most existing work focuses on learning a ranking function~\cite{diaz2012real,chaudhuri2015online,yun2014ranking},
we instead focus on direct principled approaches that are efficient (sublinear in the number of nodes), can be directly computed in real-time, easily parallelizable, and naturally amenable for online real-time ranking in the streaming setting.
This work introduces the general notion of closing higher-order motifs and based on this notion we 
develop direct ranking techniques that are efficient for real-time online ranking and prediction.
Compared to similar techniques that can be used for this setting such as Common Neighbors and methods based on it (\eg, Jaccard similarity), the proposed techniques are fundamentally more powerful as they naturally generalize over these existing techniques that are all based on closing triangles (a lower-order motif).
The proposed notion of higher-order motif closure can serve as a basis for studying and developing better ranking (and prediction) methods based on the higher-order motif closures.

\section{Closing Higher-Order Motifs} \label{sec:approach}
We first introduce the notion of a higher-order \emph{network motif closure} that lies at the heart of this work.
\begin{Definition}[Motif Closure]\label{def:motif-closure}
A node pair $(i,j)$ is said to \emph{close} a network motif $H$ iff adding an edge $(i,j)$ to $E$ \emph{closes} an instance $F \in I_G^{\prime}(H)$ of motif $H$ where $G^{\prime} = (V, E\cup\{(i,j)\})$ and $I_G^{\prime}(H)$ is the set of unique instances of motif $H$ in $G^{\prime}$.
\end{Definition}

Figure~\ref{fig:network-motif-operators} provides a few examples of higher-order motif closures.
The edge $(i,j)$ shown as a dotted line in Figure~\ref{fig:network-motif-operators} closes each motif.
For instance, the edge between node $i$ and $j$ in the rightmost motif in Figure~\ref{fig:network-motif-operators} closes a 4-clique.
We now formally introduce the frequency of higher-order motif closures for a node pair $(i,j)$ as follows:

\begin{Definition}[Higher-Order Motif Closure Frequency]\label{def:closing-higher-order-motifs}
Let $G^{\prime} = (V, E^{\prime})$ where $E^{\prime} = E\cup\{(i,j)\}$ and let $I_G^{\prime}(H)$ be the set of unique instances of motif $H$ in $G^{\prime}$.
Then the frequency of closing a higher-order motif $H$ between node $i$ and $j$ is:
\begin{equation}
W_{ij}=\sum_{F\in I_{G^{\prime}}(H)} \!\!\phantom{\Big(}\mathbb{I}\big(\{i,j\}\in E^{\prime}(F)\big)
\end{equation}\noindent
where $W_{ij}$ is equal to the number of unique instances of $H$ that contain nodes $\{i,j\}\subset V(G^{\prime})$ as an edge.
\end{Definition}

\begin{table*}[h]
\centering
\setlength{\tabcolsep}{3.7pt}
\renewcommand{\arraystretch}{1.2}
\scriptsize
\caption{
Coverage ($\downarrow$) results for the ranking methods based on closing higher-order network motifs.
Lower is better.
}
\vspace{-3mm}
\label{table:results-coverage-10perc}
\small
\fontsize{8.0}{9.0}\selectfont
\npdecimalsign{.}
\nprounddigits{3}
\begin{tabularx}{1.0\linewidth}{@{}
rccccccc c c cccccccc@{}}
\toprule
&
\multicolumn{1}{l}{\rotatebox{80}{\textsf{bn-mouse}}}  &
\multicolumn{1}{l}{\rotatebox{80}{\textsf{bio-DM-LC}}}  &
\multicolumn{1}{l}{\rotatebox{80}{\textsf{bio-CE-HT}}}  &
\multicolumn{1}{l}{\rotatebox{80}{\textsf{bio-DM-HT}}}  &
\multicolumn{1}{l}{\rotatebox{80}{\textsf{ia-reality}}}  &
\multicolumn{1}{l}{\rotatebox{80}{\textsf{web-polblogs}}}  &
\multicolumn{1}{l}{\rotatebox{80}{\textsf{biogrid-worm}}}  &
\multicolumn{1}{l}{\rotatebox{80}{\textsf{biogrid-plant}}}  &
\multicolumn{1}{l}{\rotatebox{80}{\textsf{biogrid-yeast}}}  &
\multicolumn{1}{l}{\rotatebox{80}{\textsf{email-dnc-corec.}}}  &
\multicolumn{1}{l}{\rotatebox{80}{\textsf{soc-advogato}}}  &
\multicolumn{1}{l}{\rotatebox{80}{\textsf{econ-wm1}}}  &
\multicolumn{1}{l}{\rotatebox{80}{\textsf{bn-macaque-rhe.}}}  &
\multicolumn{1}{l}{\rotatebox{80}{\textsf{road-minnesota}}}  &
\multicolumn{1}{l}{\rotatebox{80}{\textsf{soc-fb-messages}}}  &
\multicolumn{1}{l}{\rotatebox{80}{\textsf{email-EU}}}  &
\multicolumn{1}{l}{\rotatebox{80}{\textsf{email-univ}}}  \\

\midrule
\textbf{4-path}  &   0.606  &   0.964  &   0.822  &   0.962  &   0.537  &   0.958  &   0.828  &   0.980  &   0.963  &   0.911  &   0.936  &   0.976  &   1  &   \textbf{0.999}  &   0.970  &   0.579  &   0.991  \\
\textbf{4-star}  &   0.637  &   0.950  &   \textbf{0.815}  &   0.944  &   \textbf{0.024}  &   0.911  &   0.972  &   0.963  &   0.945  &   0.998  &   0.985  &   0.915  &   \textbf{0.088}  &   \textbf{0.999}  &   0.953  &   \textbf{0.245}  &   0.986  \\
\textbf{4-cycle}  &   \textbf{0.300}  &   \textbf{0.375}  &   0.922  &   \textbf{0.645}  &   0.457  &   0.942  &   \textbf{0.542}  &   \textbf{0.483}  &   0.869  &   0.801  &   0.952  &   0.942  &   0.995  &   1  &   0.971  &   0.366  &   0.897  \\
\textbf{4-tailed-triangle}  &   1  &   0.910  &   1  &   1  &   0.451  &   0.967  &   0.654  &   0.756  &   0.986  &   0.893  &   0.997  &   0.964  &   1  &   1  &   0.981  &   0.404  &   0.992  \\
\textbf{4-chordal-cycle}  &   1  &   0.913  &   1  &   0.913  &   0.936  &   \textbf{0.613}  &   1  &   1  &   \textbf{0.228}  &   \textbf{0.211}  &   \textbf{0.217}  &   0.796  &   0.965  &   1  &   1  &   0.620  &   \textbf{0.390}  \\
\textbf{4-clique}  &   1  &   1  &   1  &   1  &   1  &   0.709  &   1  &   1  &   0.322  &   0.216  &   0.315  &   \textbf{0.521}  &   0.989  &   1  &   0.902  &   0.62  &   0.409  \\
\midrule
\textbf{CN}  &   1  &   0.732  &   1  &   0.826  &   0.652  &   0.705  &   0.752  &   0.739  &   0.788  &   0.224  &   0.782  &   0.894  &   0.972  &   1  &   \textbf{0.864}  &   0.461  &   0.766  \\
\textbf{Jaccard Sim.}  &   1  &   0.732  &   1  &   0.826  &   0.658  &   0.707  &   0.752  &   0.739  &   0.824  &   0.225  &   0.782  &   0.902  &   0.960  &   1  &   0.885  &   0.462  &   0.762  \\
\textbf{Adamic/Adar}  &   1  &   0.732  &   1  &   0.826  &   0.653  &   0.706  &   0.747  &   0.739  &   0.823  &   0.218  &   0.783  &   0.905  &   0.923  &   1  &   0.866  &   0.462  &   0.761  \\

\bottomrule
\end{tabularx}
\npnoround
\end{table*}

We provide a simple routine in Algorithm~\ref{alg:higher-order-motif-closures} for computing the weight $W_{ij}$ representing the frequency of closing motif $H$ between node $i$ and $j$.
The approach has two simple steps.
First, given an arbitrary node pair $(i,j)$, a motif $H$ of interest, and the current graph $G=(V,E)$, we simply add the node pair $(i,j)$ as an edge by setting $E^{\prime} \leftarrow E \cup \{(i,j)\}$ and $G^{\prime} = (V,E^{\prime})$ (Alg.~\ref{alg:higher-order-motif-closures} Line~\ref{algline:add-fake-edge}).\footnote{Note that if edges are arriving continuously over time in a streaming fashion, then we may also encounter a node $i$ (or $j$) such that $i \not\in V$. In this case, we also set $V^{\prime} \leftarrow V\cup\{i\}$ and $G^{\prime} = (V^{\prime}, E^{\prime})$.}
As an aside, this can be performed \emph{implicitly} without any additional work.
However, it is shown in Algorithm~\ref{alg:higher-order-motif-closures} since after adding $(i,j)$ to the edge set, we can use the fastest known algorithm for counting the occurrences of motif (induced subgraph/graphlet) $H$ between node $i$ and $j$ in $G^{\prime}$.
Nevertheless, we can always modify the best known algorithm~\cite{pgd} so that it \emph{implicitly} treats the pair of nodes $(i,j)$ given as input as being connected for the sake of determining the number of instances of $H$ that would be closed if $(i,j)$ were to really exist as an edge in $G$.
Second, we compute the number of instances of motif $H$ that contain nodes $i$ and $j$ in $G^{\prime}$ (Alg.~\ref{alg:higher-order-motif-closures} Line~\ref{algline:compute-frequency-of-H}).
Given a set $\mathcal{Y}=\{y_1,y_2,\ldots,y_j,\ldots\}$ of nodes (items, ads, songs, friends) to be ranked, Algorithm~\ref{alg:higher-order-motif-closures} can be used to obtain $W_{ij}=f(x_i,y_j)$, $\forall j=1,\ldots,|\mathcal{Y}|$.

{
\algblockdefx[parallel]{parfor}{endpar}[1][]{$\textbf{parallel for}$ #1 $\textbf{do}$}{$\textbf{end parallel}$}
\algrenewcommand{\alglinenumber}[1]{\fontsize{7.0}{8.0}\selectfont#1\;\;}
\begin{figure}[h!]
\vspace{-3mm}
\begin{center}
\begin{algorithm}[H]
\caption{\,
Higher-Order Motif Closures
}
\label{alg:higher-order-motif-closures}
\begin{spacing}{1.15}
\small
\begin{algorithmic}[1]
\smallskip
\Require a graph $G=(V,E)$, node pair $(i,j)$, and network motif/graphlet $H$ 
\Ensure the frequency $W_{ij}$ of motif closures of $H$ for nodes $i$ and $j$
\smallskip

\State Set $E^{\prime} \leftarrow E \cup \{(i,j)\}$ and $G^{\prime} = (V,E^{\prime})$ \label{algline:add-fake-edge}

\State Use fast algorithm~\protect\cite{pgd,rossi17graphlet-est} to compute $W_{ij} = $ \# of occurrences of motif $H$ between node $i$ and $j$ in $G^{\prime}$ 
\label{algline:compute-frequency-of-H}
\smallskip
\end{algorithmic}
\end{spacing}
\end{algorithm}
\end{center}
\vspace{-5mm}
\end{figure}
}

\begin{figure*}[h!]
\vspace{-2mm}
\centering
\hspace{-6mm}
\subfigure{
\includegraphics[width=0.27\linewidth]{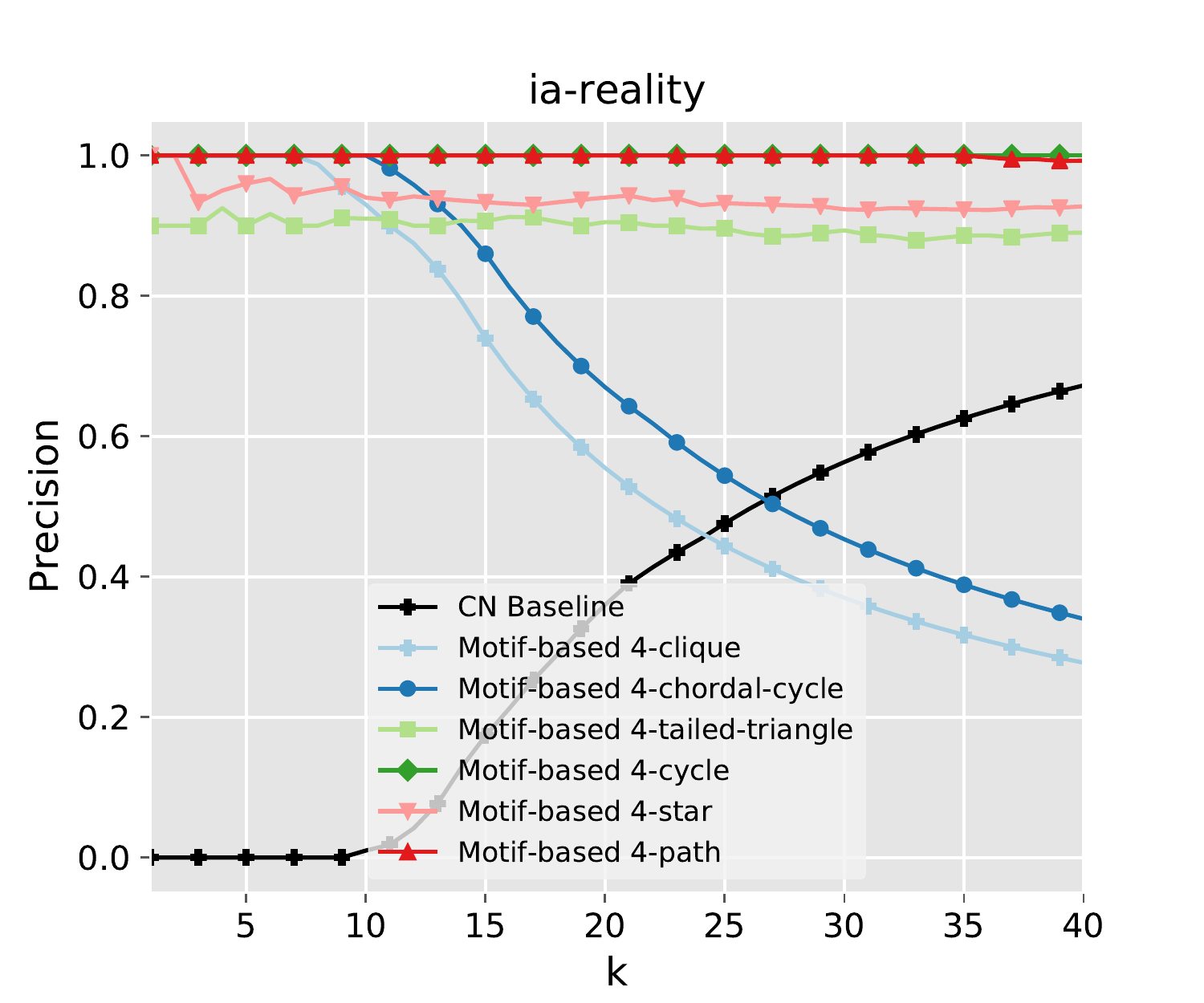}
}
\hspace{-6.5mm}
\subfigure{
\includegraphics[width=0.27\linewidth]{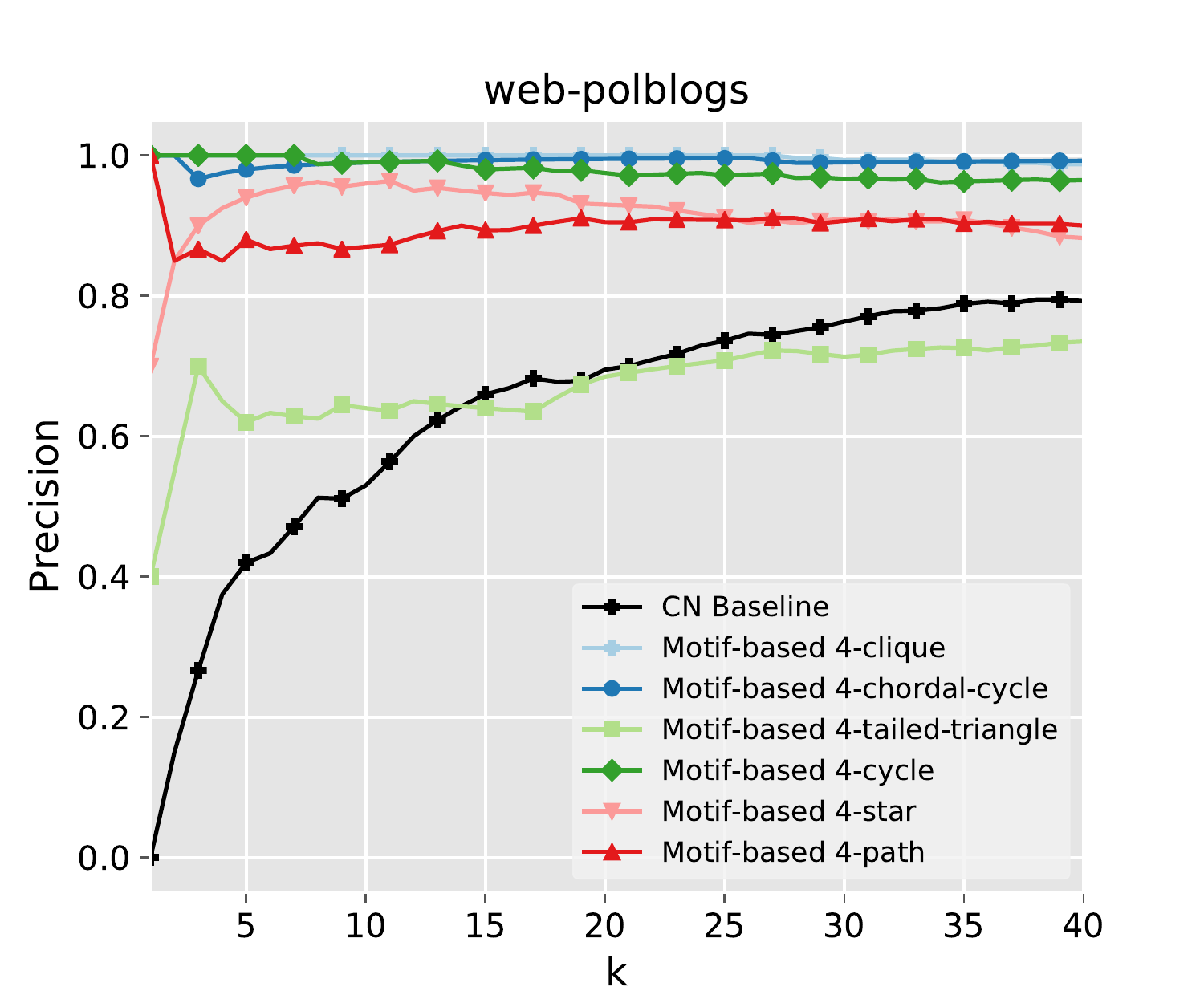}
}
\hspace{-6.5mm}
\subfigure{
\includegraphics[width=0.27\linewidth]{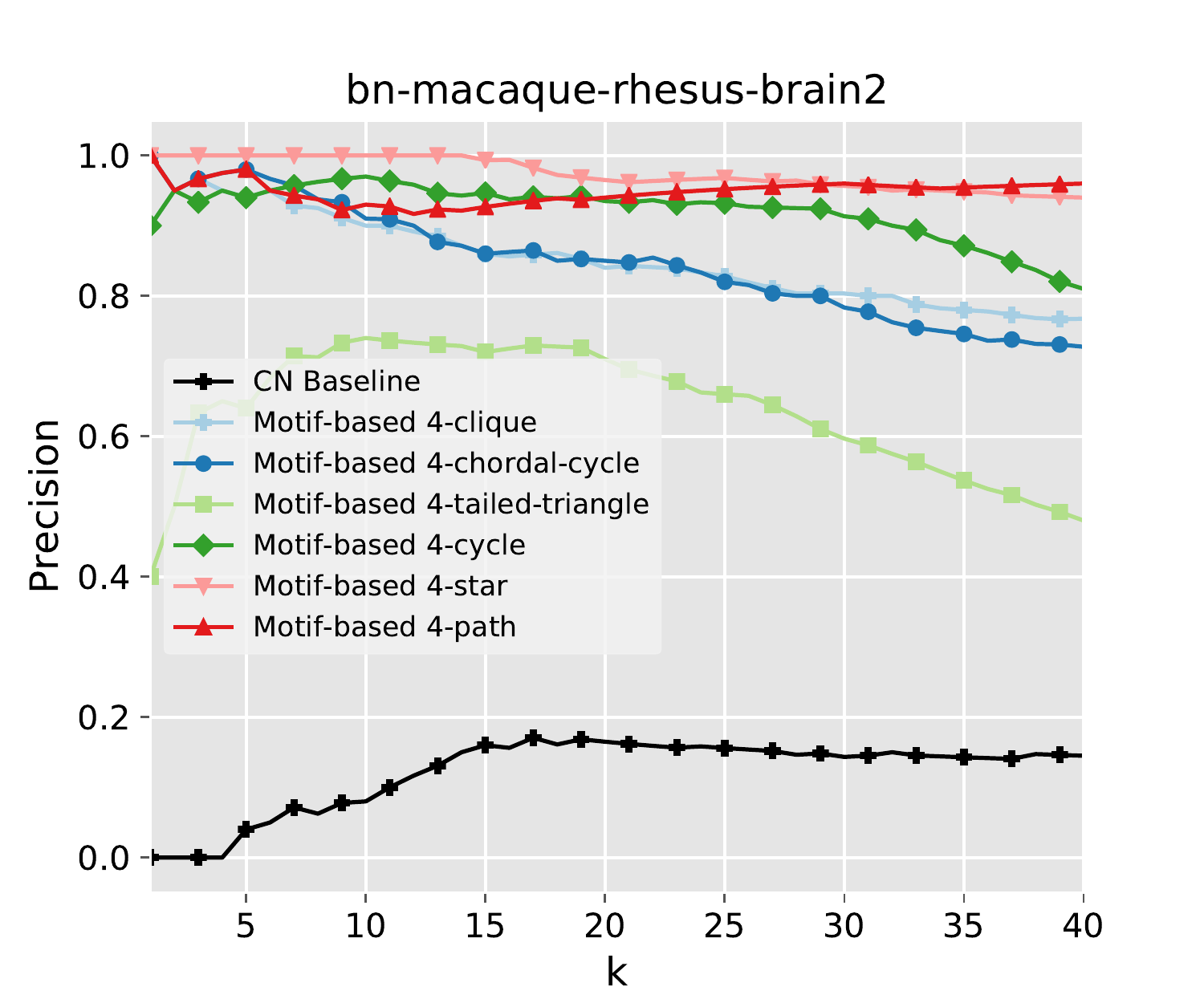}
}
\hspace{-6.5mm}
\subfigure{
\includegraphics[width=0.27\linewidth]{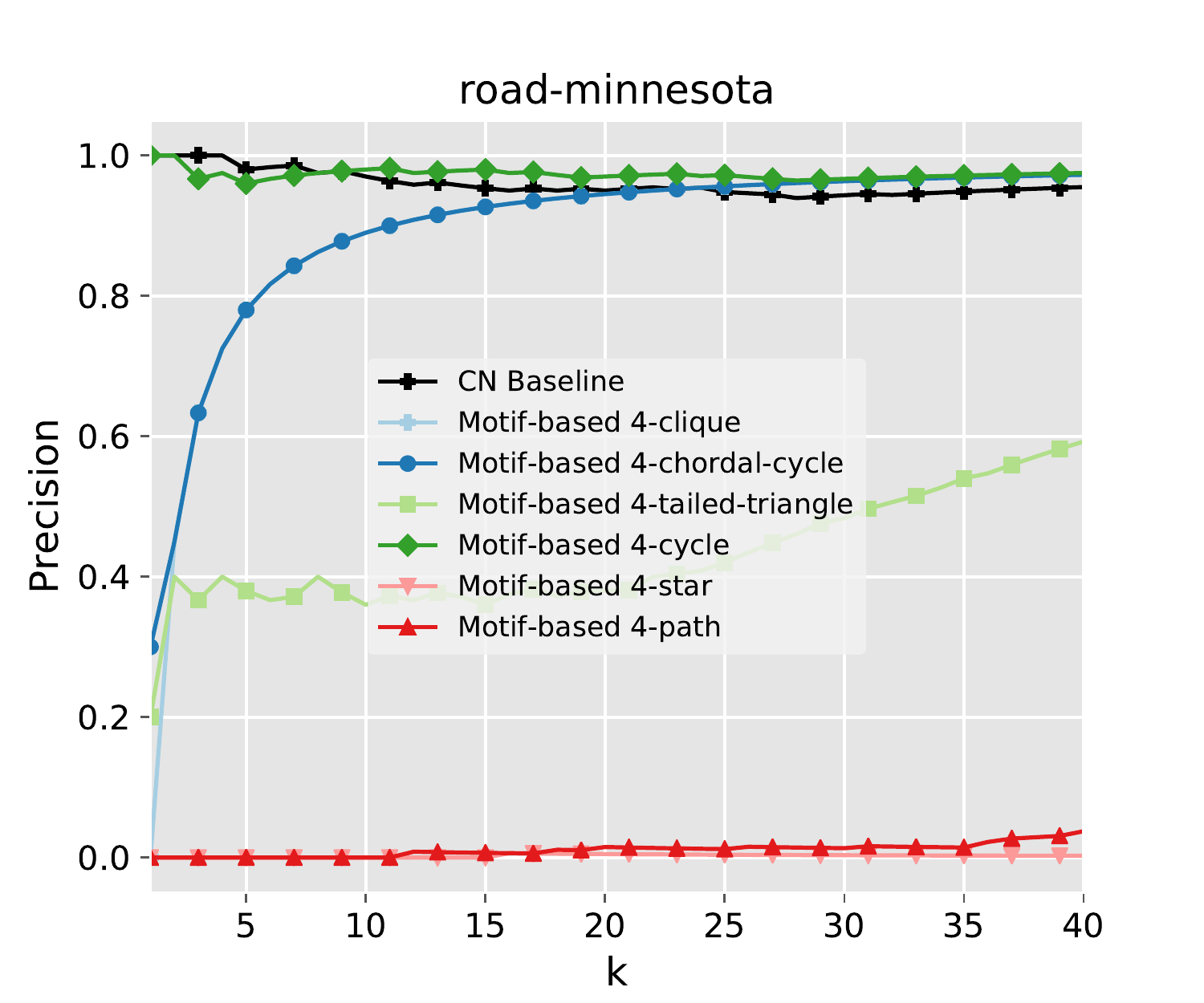}
}
\hspace{-6mm}

\vspace{-4mm}
\caption{Precision at $\mathbf{k=1,\ldots,40}$ for different network motif closure rankings.
}
\label{fig:prec-at-k-closing-diff-higher-order-motifs-top40}
\end{figure*}

\subsubsection*{Extending Other Measures using Motif Closure}
Given two nodes $i$ and $j$, Common Neighbor-based methods are those that use the quantity $|\Gamma_i\cap \Gamma_j|$ where $\Gamma_i$ and $\Gamma_j$ are the set of neighbors for node $i$ and $j$, respectively.
Common neighbors is simply $W_{ij} = |\Gamma_i\cap \Gamma_j|$ where $W_{ij}$ represents the number of \emph{potential triangles} that \emph{would be closed} if there were an edge between $i$ and $j$.
The notion of ``closing'' triangles lies at the heart of many other existing methods that are based on $|\Gamma_i \cap \Gamma_j|$ such as 
Jaccard similarity, Adamic/Adar (AA), among others.
All of these methods can be viewed as extensions of Common Neighbors with some form of normalization, \eg, Jaccard similarity is $W_{ij} = |\Gamma_i \cap \Gamma_j| / |\Gamma_i \cup \Gamma_j|$.
Extending the proposed higher-order motif-based link ranking and prediction techniques is left for future work. This includes extending the notion of ``closing'' higher-order network motifs for other measures such as Jaccard similarity, Adamic/Adar, among any others where the notion of \emph{closing triangles} can be replaced with the notion of
\emph{closing a higher-order motif} introduced in this work.

\section{Experiments} \label{sec:exp}
The experiments are designed to evaluate the effectiveness of the proposed methods that are based on the notion of ``closing'' higher-order network motifs.
These methods go beyond closing simple triangles.
To ensure the significance and generality of our findings (as much as possible), we evaluate the proposed methods using a wide variety of networks from different application domains.
All data was obtained from NetworkRepository~\cite{nr}.

We compare the proposed higher-order motif closure methods against CN-based methods (CN, Jaccard similarity, Adamic/Adar) since these are all based on closing triangles also have the same desired properties as the 
higher-order motif closure methods described in this paper.
In this work, we only investigate the most basic and fundamental higher-order motif closures.
Developing more sophisticated higher-order ranking measures based on these fundamental motif closures is left for future work.
However, we did run a few experiments using an extended higher-order Jaccard similarity (one for each motif closure, giving 6 total for 4-node motifs) and higher-order Adamic/Adar ranking measures, again giving 6 new rankings total.
Since each variant provides 6 additional rankings, the results were removed for brevity, but in some cases performed better than the most basic motif closures introduced in this paper.
As such, the proposed notion of higher-order motif closures serve as fundamental building blocks for developing better higher-order ranking and prediction methods.

Unless otherwise mentioned, we hold-out 10\% of the observed node pairs and randomly sample the same number of negative node pairs. We then use the methods to obtain a ranking of the node pairs in this set.\footnote{In recommender systems, the set of node pairs to be ranked is actually a smaller set of ``relevant items'' $\mathcal{Y}_{i}=\{y_1,\ldots,y_j,\ldots\} \subset \mathcal{Y}$ for a user $i$. Nevertheless, this can also be viewed as a ranking of node pairs where user $i$ is fixed.} 
Recall the proposed techniques do not require learning a sophisticated model nor do they require training data.
As such, the notion of motif closure proposed in this work can be used in a real-time streaming fashion and has many obvious advantages to more sophisticated model-based approaches.
Mean Average Precision (MAP) results are provided in Table~\ref{table:results-MAP-10perc} whereas coverage is provided in Table~\ref{table:results-coverage-10perc}.

\begin{Result}\label{res:motif-closure-outperforms-triangle-closure}
Ranking based on closing higher-order motifs outperforms other direct methods that are based on closing triangles.
\end{Result}
In nearly all cases, the higher-order motif closures achieve better precision and coverage than techniques based on closing lower-order triangles.

\begin{Result}\label{res:consistency-of-best-performing-closure}
The best performing motif closure is consistent across different evaluation measures.
The motif closure that achieves the best \emph{precision} (Table~\ref{table:results-MAP-10perc}) is typically the same motif that achieves the best \emph{coverage} (Table~\ref{table:results-coverage-10perc}). 
\end{Result}

\begin{Result}\label{res:best-performing-closure}
There is no single higher-order motif closure that performs best for all graphs.
The best motif depends highly on the structural characteristics of the graph and its domain (biological vs. social network) as shown in Table~\ref{table:results-MAP-10perc} and Table~\ref{table:results-coverage-10perc}.
\end{Result}

In Table~\ref{table:results-MAP-10perc}-\ref{table:results-coverage-10perc}, biological and brain networks achieve best performance using the ranking given by 4-cycle and 4-star closures.
This also holds true for the interaction (ia-reality) and road network investigated.
The 4-star and 4-cycle motif closures are more sparse compared to the 4-chordal-cycle (paw motif) and 4-clique motif closure.
In the web graph, economic, and social networks, both the 4-chordal-cycle (diamond motif closure) and 4-clique motif closure achieves significantly better performance than the other motif closures.
Notice that both these motif closures are composed of two or more triangles and thus can be seen as a stronger triadic closure motif.
The 4-path, 4-tailed-triangle, and triangle (CN) motif closures did not perform the best in any of the graphs investigated.
That is, there were always a higher-order motif closure with better performance as shown in Table~\ref{table:results-MAP-10perc} and Table~\ref{table:results-coverage-10perc}.
In Figure~\ref{fig:prec-at-k-closing-diff-higher-order-motifs-top40}, we also show the precision at $k=1,\ldots,40$ for closing different higher-order network motifs.
In nearly all cases, the rankings given by the 4-node motif closures are better than the lower-order CN approach that is based on closing triangles.

\begin{table}[t!]
\vspace{-1mm}
\centering
\setlength{\tabcolsep}{2.15pt}
\renewcommand{\arraystretch}{1.2}
\scriptsize
\caption{Robustness results (MAP). See text for discussion.}
\vspace{-3mm}
\label{table:results-MAP-robustness}
\small
\fontsize{8.0}{9.0}\selectfont
\npdecimalsign{.}
\nprounddigits{3}
\begin{tabularx}{1.0\linewidth}{@{}
rccccccccccccccc@{}}
\toprule
&
\multicolumn{1}{l}{\rotatebox{80}{\textsf{bn-mouse}}}  &
\multicolumn{1}{l}{\rotatebox{80}{\textsf{bio-DM-LC}}}  &
\multicolumn{1}{l}{\rotatebox{80}{\textsf{bio-CE-HT}}}  &
\multicolumn{1}{l}{\rotatebox{80}{\textsf{bio-DM-HT}}}  &
\multicolumn{1}{l}{\rotatebox{80}{\textsf{ia-reality}}}  &
\multicolumn{1}{l}{\rotatebox{80}{\textsf{web-polblogs}}}  &
\multicolumn{1}{l}{\rotatebox{80}{\textsf{biogrid-worm}}}  &
\multicolumn{1}{l}{\rotatebox{80}{\textsf{biogrid-plant}}}  \\
\midrule

\textbf{4-path}  &   0.764  &   0.698  &   0.495  &   0.582  &   0.774  &   0.812  &   \textbf{0.899}  &   0.847  \\
\textbf{4-star}  &   \textbf{0.868}  &   0.781  &   \textbf{0.52}  &   0.645  &   \textbf{0.920}  &   0.81  &   0.894  &   0.836  \\
\textbf{4-cycle}  &   0.763  &   \textbf{0.915}  &   0.482  &   \textbf{0.871}  &   0.665  &   0.855  &   0.863  &   \textbf{0.872}  \\
\textbf{4-tailed-triangle}  &   0.684  &   0.700  &   0.444  &   0.755  &   0.725  &   0.708  &   0.765  &   0.658  \\
\textbf{4-chordal-cycle}  &   0.781  &   0.816  &   0.440  &   0.811  &   0.237  &   0.942  &   0.812  &   0.776  \\
\textbf{4-clique}  &   0.781  &   0.825  &   0.440  &   0.809  &   0.205  &   \textbf{0.944}  &   0.807  &   0.761  \\
\midrule
\textbf{CN}  &   0.686  &   0.800  &   0.458  &   0.801  &   0.283  &   0.806  &   0.791  &   0.819  \\
\textbf{Jaccard Sim.}  &   0.693  &   0.809  &   0.461  &   0.803  &   0.357  &   0.910  &   0.808  &   0.828  \\
\textbf{Adamic/Adar}  &   0.704  &   0.809  &   0.462  &   0.803  &   0.375  &   0.912  &   0.808  &   0.831  \\

\bottomrule
\end{tabularx}
\npnoround
\vspace{-1mm}
\end{table}

\subsubsection*{Robustness of Ranking from Higher-Order Motif Closures}
In addition, we investigate the robustness of the higher-order motif closures to noise in the graph, \ie, random link additions.
To understand the robustness of the motif closure methods for graphs with noisy and spurious links, we select pairs of nodes uniformly at random that are not linked in $G$ and create a link between each pair.
In this set of experiments, we sample $|E|/2$ node pairs (negative/unobserved edges) and add them to $G$.
Results are shown in Table~\ref{table:results-MAP-robustness}.
Due to space, we show only a subset of the networks used in Table~\ref{table:results-MAP-10perc}-\ref{table:results-coverage-10perc}.
\begin{Result}\label{res:robustness}
Robustness of the ranking by higher-order motif closures is comparable, and slightly better than techniques based on closing triangles.
\end{Result}

\begin{figure}[h!]
\hspace{-6mm}\includegraphics[width=0.74\linewidth]{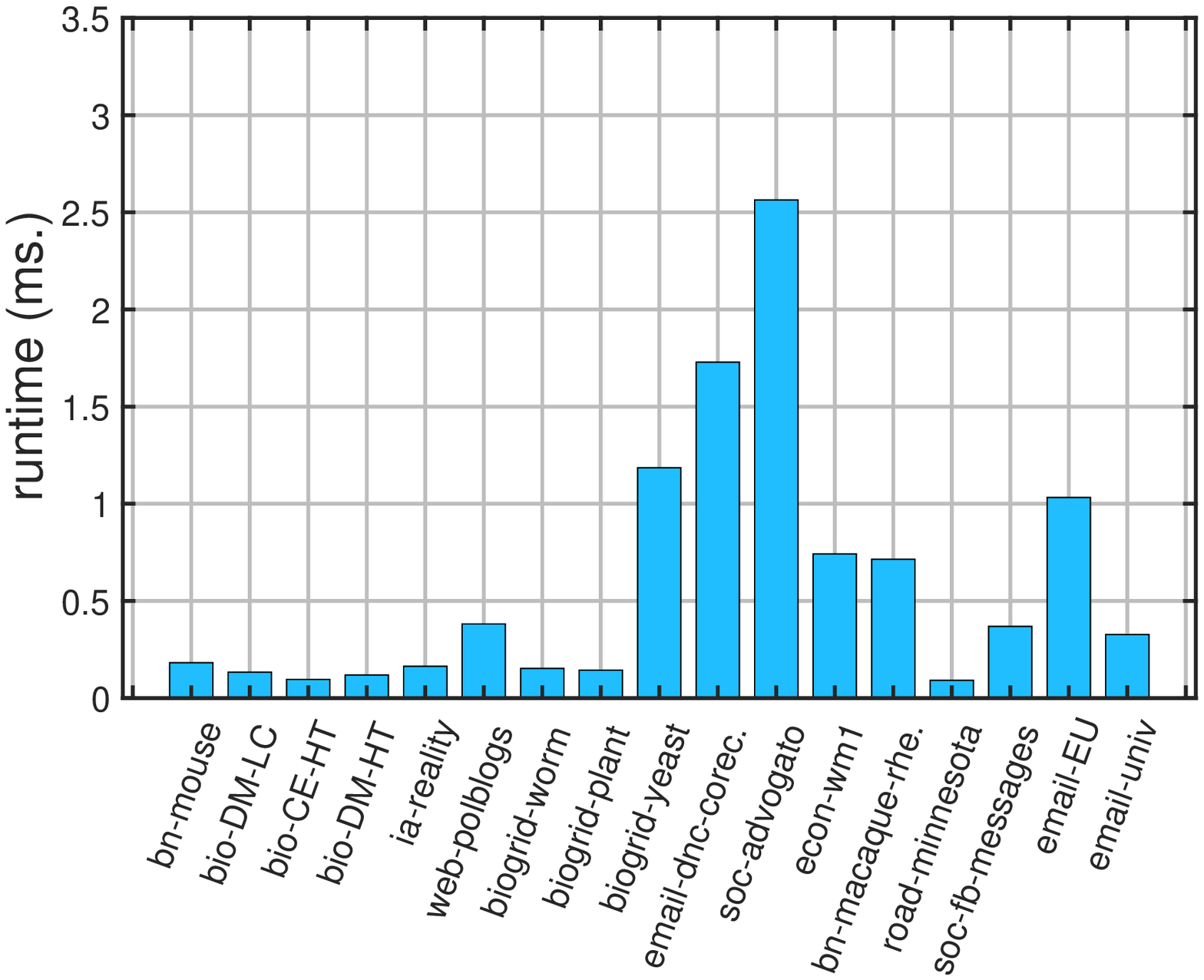}

\vspace{-2.5mm}
\caption{Average runtime in milliseconds to compute all motif closures for each node pair.}
\label{fig:runtime-ms-per-edge-motif-closure}
\vspace{-1mm}
\end{figure}

\subsubsection*{Runtime performance}
We report the average runtime in milliseconds to compute all motif closures for each node pair in $G$.
For most graphs, it takes \emph{less than a millisecond} on average as shown in Figure~\ref{fig:runtime-ms-per-edge-motif-closure} and therefore is fast for large-scale ranking problems.

\begin{Result}\label{res:motif-closure-runtime}
For any 4-node motif $H$, counting the number of motif closures $W_{ij}$ that would arise if $(i,j)$ was added to $G$ is fast taking less than a millisecond on average (across all graphs).
\end{Result}

The runtime can be significantly improved for certain problem settings:
Suppose we are interested in only the top-$k$ most relevant node pairs (or items for a user $i$) given by a ranking from an arbitrary motif closure for motif $H$,
then for possibly many such node pairs, we can avoid computing $W_{ij}$ (\ie, \# of instances of motif $H$ in $G$ that would be closed if the node pair $(i,j)$ actually existed/observed in $G$) altogether by first deriving an upper bound $\textrm{UB}$ of $W_{ij}$ in $o(1)$ constant time and only computing $W_{ij}$ if 
$\textrm{UB} > \delta$ where $\delta$ is the weight of the node pair in the top-$k$ ranking with minimum weight (the node pair with rank $k$).
Since otherwise we know $W_{ij}$ is not large enough to beat the node pair with the $k$-th largest weight.

\section{Conclusion} \label{sec:conc}
This work proposed the notion of \emph{motif closure} and described higher-order ranking and link prediction techniques based on the notion of \emph{closing higher-order network motifs}.
Such techniques were shown to be effective for online real-time ranking (and prediction) as they often outperformed a number of baselines that are based on closing triangles.
Future work will investigate using the notion of \emph{closing higher-order motifs} to extend other techniques such as a higher-order Jaccard similarity or higher-order Adamic/Adar measures based on closing higher-order network motifs such as 4-cliques, 4-cycles, among others.

\small
\balance
\bibliographystyle{ACM-Reference-Format}
\bibliography{paper}


\begin{thebibliography}{00}

%%% the Web address.

\ifx \showCODEN    \undefined \def \showCODEN     #1{\unskip}     \fi
\ifx \showDOI      \undefined \def \showDOI       #1{#1}\fi
\ifx \showISBNx    \undefined \def \showISBNx     #1{\unskip}     \fi
\ifx \showISBNxiii \undefined \def \showISBNxiii  #1{\unskip}     \fi
\ifx \showISSN     \undefined \def \showISSN      #1{\unskip}     \fi
\ifx \showLCCN     \undefined \def \showLCCN      #1{\unskip}     \fi
\ifx \shownote     \undefined \def \shownote      #1{#1}          \fi
\ifx \showarticletitle \undefined \def \showarticletitle #1{#1}   \fi
\ifx \showURL      \undefined \def \showURL       {\relax}        \fi
% invisible to TeX
\providecommand\bibfield[2]{#2}
\providecommand\bibinfo[2]{#2}
\providecommand\natexlab[1]{#1}
\providecommand\showeprint[2][]{arXiv:#2}

\bibitem[\protect\citeauthoryear{Ahmed, Neville, Rossi, and Duffield}{Ahmed
  et~al\mbox{.}}{2015}]%
        {pgd}
\bibfield{author}{\bibinfo{person}{Nesreen~K. Ahmed}, \bibinfo{person}{Jennifer
  Neville}, \bibinfo{person}{Ryan~A. Rossi}, {and} \bibinfo{person}{Nick
  Duffield}.} \bibinfo{year}{2015}\natexlab{}.
\newblock \showarticletitle{Efficient Graphlet Counting for Large Networks}. In
  \bibinfo{booktitle}{{\em ICDM}}. \bibinfo{pages}{10}.
\newblock


\bibitem[\protect\citeauthoryear{Berry and Browne}{Berry and Browne}{2005}]%
        {berry2005understanding}
\bibfield{author}{\bibinfo{person}{Michael~W Berry} {and}
  \bibinfo{person}{Murray Browne}.} \bibinfo{year}{2005}\natexlab{}.
\newblock \bibinfo{booktitle}{{\em Understanding Search Engines: Mathematical
  Modeling and Text Retrieval}}. Vol.~\bibinfo{volume}{17}.
\newblock \bibinfo{publisher}{SIAM}.
\newblock


\bibitem[\protect\citeauthoryear{Chaudhuri and Tewari}{Chaudhuri and
  Tewari}{2015}]%
        {chaudhuri2015online}
\bibfield{author}{\bibinfo{person}{Sougata Chaudhuri} {and}
  \bibinfo{person}{Ambuj Tewari}.} \bibinfo{year}{2015}\natexlab{}.
\newblock \showarticletitle{Online ranking with top-1 feedback}. In
  \bibinfo{booktitle}{{\em Artificial Intelligence and Statistics}}.
  \bibinfo{pages}{129--137}.
\newblock


\bibitem[\protect\citeauthoryear{Deshpande and Karypis}{Deshpande and
  Karypis}{2004}]%
        {deshpande2004item}
\bibfield{author}{\bibinfo{person}{Mukund Deshpande} {and}
  \bibinfo{person}{George Karypis}.} \bibinfo{year}{2004}\natexlab{}.
\newblock \showarticletitle{Item-based top-n recommendation algorithms}.
\newblock \bibinfo{journal}{{\em TOIS\/}} \bibinfo{volume}{22},
  \bibinfo{number}{1} (\bibinfo{year}{2004}), \bibinfo{pages}{143--177}.
\newblock


\bibitem[\protect\citeauthoryear{Diaz-Aviles, Drumond, Schmidt-Thieme, and
  Nejdl}{Diaz-Aviles et~al\mbox{.}}{2012}]%
        {diaz2012real}
\bibfield{author}{\bibinfo{person}{Ernesto Diaz-Aviles}, \bibinfo{person}{Lucas
  Drumond}, \bibinfo{person}{Lars Schmidt-Thieme}, {and}
  \bibinfo{person}{Wolfgang Nejdl}.} \bibinfo{year}{2012}\natexlab{}.
\newblock \showarticletitle{Real-time top-n recommendation in social streams}.
  In \bibinfo{booktitle}{{\em RecSys}}. ACM, \bibinfo{pages}{59--66}.
\newblock


\bibitem[\protect\citeauthoryear{Liben-Nowell and Kleinberg}{Liben-Nowell and
  Kleinberg}{2007}]%
        {liben2007link}
\bibfield{author}{\bibinfo{person}{David Liben-Nowell} {and}
  \bibinfo{person}{Jon Kleinberg}.} \bibinfo{year}{2007}\natexlab{}.
\newblock \showarticletitle{The link-prediction problem for social networks}.
\newblock \bibinfo{journal}{{\em JASIST\/}} \bibinfo{volume}{58},
  \bibinfo{number}{7} (\bibinfo{year}{2007}), \bibinfo{pages}{1019--1031}.
\newblock


\bibitem[\protect\citeauthoryear{Manning, Raghavan, and Sch{\"u}tze}{Manning
  et~al\mbox{.}}{2010}]%
        {manning2010introduction}
\bibfield{author}{\bibinfo{person}{Christopher Manning},
  \bibinfo{person}{Prabhakar Raghavan}, {and} \bibinfo{person}{Hinrich
  Sch{\"u}tze}.} \bibinfo{year}{2010}\natexlab{}.
\newblock \showarticletitle{Introduction to information retrieval}.
\newblock \bibinfo{journal}{{\em Nat. Lang. Eng.\/}} \bibinfo{volume}{16},
  \bibinfo{number}{1} (\bibinfo{year}{2010}), \bibinfo{pages}{100--103}.
\newblock


\bibitem[\protect\citeauthoryear{Rossi and Ahmed}{Rossi and Ahmed}{2015}]%
        {nr}
\bibfield{author}{\bibinfo{person}{Ryan~A. Rossi} {and}
  \bibinfo{person}{Nesreen~K. Ahmed}.} \bibinfo{year}{2015}\natexlab{}.
\newblock \showarticletitle{The Network Data Repository with Interactive Graph
  Analytics and Visualization}. In \bibinfo{booktitle}{{\em AAAI}}.
  \bibinfo{pages}{4292--4293}.
\newblock
\showURL{%
\url{http://networkrepository.com}}


\bibitem[\protect\citeauthoryear{Rossi, McDowell, Aha, and Neville}{Rossi
  et~al\mbox{.}}{2012}]%
        {rossi12jair}
\bibfield{author}{\bibinfo{person}{Ryan~A. Rossi}, \bibinfo{person}{Luke~K.
  McDowell}, \bibinfo{person}{David~W. Aha}, {and} \bibinfo{person}{Jennifer
  Neville}.} \bibinfo{year}{2012}\natexlab{}.
\newblock \showarticletitle{Transforming graph data for statistical relational
  learning}.
\newblock \bibinfo{journal}{{\em JAIR\/}} (\bibinfo{year}{2012}),
  \bibinfo{pages}{363--441}.
\newblock


\bibitem[\protect\citeauthoryear{Rossi, Zhou, and Ahmed}{Rossi
  et~al\mbox{.}}{2018}]%
        {rossi17graphlet-est}
\bibfield{author}{\bibinfo{person}{Ryan~A. Rossi}, \bibinfo{person}{Rong Zhou},
  {and} \bibinfo{person}{Nesreen~K. Ahmed}.} \bibinfo{year}{2018}\natexlab{}.
\newblock \showarticletitle{Estimation of Graphlet Counts in Massive Networks}.
  In \bibinfo{booktitle}{{\em TNNLS}}. \bibinfo{pages}{44--57}.
\newblock


\bibitem[\protect\citeauthoryear{Tsitsulin, Mottin, Karras, and
  M{\"u}ller}{Tsitsulin et~al\mbox{.}}{2018}]%
        {tsitsulin2018verse}
\bibfield{author}{\bibinfo{person}{Anton Tsitsulin}, \bibinfo{person}{Davide
  Mottin}, \bibinfo{person}{Panagiotis Karras}, {and} \bibinfo{person}{Emmanuel
  M{\"u}ller}.} \bibinfo{year}{2018}\natexlab{}.
\newblock \showarticletitle{Verse: Versatile graph embeddings from similarity
  measures}. In \bibinfo{booktitle}{{\em WWW}}. \bibinfo{pages}{539--548}.
\newblock


\bibitem[\protect\citeauthoryear{Yun, Raman, and Vishwanathan}{Yun
  et~al\mbox{.}}{2014}]%
        {yun2014ranking}
\bibfield{author}{\bibinfo{person}{Hyokun Yun}, \bibinfo{person}{Parameswaran
  Raman}, {and} \bibinfo{person}{S Vishwanathan}.}
  \bibinfo{year}{2014}\natexlab{}.
\newblock \showarticletitle{Ranking via robust binary classification}. In
  \bibinfo{booktitle}{{\em NIPS}}. \bibinfo{pages}{2582--2590}.
\newblock


\end{thebibliography}

\end{document}